\documentclass[11pt, a4paper, logo, copyright, nonumbering]{cls/Sci}
\usepackage[numbers, sort&compress, square]{natbib}
\usepackage{dblfloatfix}
\usepackage{ulem}
\usepackage{caption}
\usepackage{dramatist}
\usepackage{wrapfig}
\usepackage{xspace}
\usepackage{pifont} %
\usepackage{multirow}

\usepackage{xltabular}
\usepackage{longtable}
\usepackage{hyperref}
\usepackage{algorithm}
\usepackage{algpseudocode}
\usepackage{amsmath}
\usepackage{amssymb}
\usepackage{amsfonts}
\usepackage{amsmath}
\usepackage{amssymb}
\usepackage{lineno}
\usepackage{multirow}
\usepackage{adjustbox}
\usepackage{cleveref}
\usepackage[bottom]{footmisc}
\usepackage{CJKutf8}
\usepackage{subfigure}
\usepackage{setspace}
\usepackage{bbding}
\usepackage{pifont}
\usepackage{dsfont}
\usepackage{array} %
\usepackage{tabularx} %
\usepackage{subfigure} %
\usepackage{xcolor} %
\usepackage{tabularx}
\usepackage{booktabs}
\usepackage{tcolorbox}
\tcbuselibrary{breakable}
\usepackage{lipsum}  %
\usepackage{multicol} %
\definecolor{chatgray}{RGB}{245,245,245}
\definecolor{chatborder}{RGB}{200,200,200}
\usepackage{float}  %

\newtcolorbox{promptbox}[1][]{
  breakable,                
  colback=chatgray,         
  colframe=chatborder,      
  coltitle=black,           
  fonttitle=\bfseries,      
  title={#1},               
  arc=3mm,                  
  boxrule=0.5pt,           
  left=2mm, right=2mm, top=2mm, bottom=2mm, 
  fontupper=\small\ttfamily 
}

\makeatletter
\def\@BTrule[#1]{%
  \ifx\longtable\undefined
    \let\@BTswitch\@BTnormal
  \else\ifx\hline\LT@hline
    \nobreak
    \let\@BTswitch\@BLTrule
  \else
     \let\@BTswitch\@BTnormal
  \fi\fi
  \global\@thisrulewidth=#1\relax
  \ifnum\@thisruleclass=\tw@\vskip\@aboverulesep\else
  \ifnum\@lastruleclass=\z@\vskip\@aboverulesep\else
  \ifnum\@lastruleclass=\@ne\vskip\doublerulesep\fi\fi\fi
  \@BTswitch}
\makeatother

\addto\extrasenglish{
    \def\sectionautorefname{Section}%

}

 {\begin{list}{}%
         {\setlength{\leftmargin}{#1}}%
         \item[]%
 }
 {\end{list}}
 
\reportnumber{001} %

\newcolumntype{Y}{>{\centering\arraybackslash}X}
\title{\centering Toward Ultra-Long-Horizon Agentic Science: Cognitive Accumulation for Machine Learning Engineering}

\author{
    Xinyu Zhu\textsuperscript{\rm 1, $\dagger$},
    Yuzhu Cai\textsuperscript{\rm 1,3, $\dagger$}, 
    Zexi Liu\textsuperscript{\rm 1, $\dagger$}, 
    Bingyang Zheng\textsuperscript{\rm 1},
    Cheng Wang\textsuperscript{\rm 1},
    
    Rui Ye\textsuperscript{\rm 1},
    Yuzhi Zhang\textsuperscript{\rm 2},
    Linfeng Zhang\textsuperscript{\rm 2},
    Weinan E\textsuperscript{\rm 1},
    
    Siheng Chen\textsuperscript{\rm 1, $\ast$},
    Yanfeng Wang\textsuperscript{\rm 1,3,$\ast$}
    \\
    \textsuperscript{\rm 1} School of Artificial Intelligence, Shanghai Jiao Tong University \\
    \textsuperscript{\rm 2} DP Technology, \textsuperscript{\rm 3} Shanghai AI Laboratory
}

\begin{document}
\begingroup
    \renewcommand{\thefootnote}{\fnsymbol{footnote}} 
    
    \footnotetext[2]{Equal contribution. Order randomized.}

    \footnotetext[1]{Corresponding author: \texttt{sihengc@sjtu.edu.cn}, \texttt{wangyanfeng622@sjtu.edu.cn}}

\endgroup

\begin{abstract}
\vspace{-1em}
The advancement of artificial intelligence toward agentic science is currently bottlenecked by the challenge of ultra-long-horizon autonomy, the ability to sustain strategic coherence and iterative correction over experimental cycles spanning days or weeks. 
While Large Language Models (LLMs) have demonstrated prowess in short-horizon reasoning, they are easily overwhelmed by execution details in the high-dimensional, delayed-feedback environments of real-world research, failing to consolidate sparse feedback into coherent long-term guidance. 
Here, we present ML-Master 2.0, an autonomous agent that masters ultra-long-horizon machine learning engineering (MLE) which is a representative microcosm of scientific discovery. 
By reframing context management as a process of cognitive accumulation, our approach introduces Hierarchical Cognitive Caching (HCC), a multi-tiered architecture inspired by computer systems that enables the structural differentiation of experience over time. 
By dynamically distilling transient execution traces into stable knowledge and cross-task wisdom, HCC allows agents to decouple immediate execution from long-term experimental strategy, effectively overcoming the scaling limits of static context windows. 
In evaluations on OpenAI's MLE-Bench under 24-hour budgets, ML-Master 2.0 achieves a state-of-the-art medal rate of 56.44\%. 
Our findings demonstrate that ultra-long-horizon autonomy provides a scalable blueprint for AI capable of autonomous exploration beyond human-precedent complexities.
\end{abstract}

\maketitle

\newpage

\begin{spacing}{0.9}
\end{spacing}

\newpage

\section{Introduction}
\label{sec:introduction}
The rapid evolution of Large Language Models (LLMs) is propelling artificial intelligence (AI) from passive assistance toward the ambitious frontier of agentic science. ~\cite{zheng2025automationautonomysurveylarge,lu2024aiscientistfullyautomated,boiko2023autonomous,ren2025towards}
This transition is already moving from theoretical speculation to experimental practice, as evidenced by a series of representative efforts, including DeepMind's AlphaEvolve~\cite{novikov2025alphaevolvecodingagentscientific}, OpenAI's FrontierScience~\cite{OpenAI2025FrontierScience}, and Google's AI co-scientist~\cite{gottweis2025aicoscientist}. 
However, shifting from isolated tasks to real-world reasearch encounters a fundamental barrier. 
Scientific Discovery is inherently a ultra-long-horizon process, characterized not by momentary acts of reasoning but by delayed feedback, high-dimensional exploration, and experimental cycles spanning days or weeks.~\cite{lu2024aiscientistfullyautomated}
This intrinsic complexity necessitates \textbf{\textit{ultra-long-horizon autonomy}}. 
It refers to the capacity to sustain strategic coherence and perform iterative correction over extended temporal scales without being overwhelmed by the accumulation of execution details. 

To address this, we situate our research within the paradigm of \textbf{\textit{AI-for-AI (AI4AI)}}, where AI systems autonomously drive the advancement of AI itself. 
Contrasted with traditional scientific discovery where validation could require specialized equipment and months of latency (e.g., in chemical experiments~\cite{boiko2023autonomous}), AI4AI offers a purely computational substrate with negligible cost and immediate feedback. This enables the accelerated experimentation required to advance {ultra-long-horizon autonomy} without the bottleneck of physical constraints.
Within this paradigm, the machine learning engineering (MLE) tasks emphasized by OpenAI's MLE-Bench~\cite{chan2024mle} emerge as its quintessential challenge. MLE-Bench is a benchmark comprising 75 real-world Kaggle machine learning competitions.
Far exceeding simple code generation, it requires agents to navigate a vast, unstructured search space through prolonged trial and error and the accumulation of experience across iterations, rather than by single-step correctness.
By treating MLE as a specific instance of \textit{ultra-long-horizon} inquiry, we aim to develop generalizable methods that enable agents to evolve their context and sustain strategic focus over the tens of hours required for genuine breakthrough.

Motivated by this perspective, we introduce \textbf{ML-Master 2.0}, an autonomous agent designed to support ultra-long-horizon MLE through cognitive accumulation. 
We posit that ultra-long-horizon autonomy is not a linear aggregation of historical context, but an evolutionary process of refinement, stabilization, and reuse. 
From a congnitive perspective, agents generate large amounts of raw \textit{experience}. 
Only a small fraction of these, once repeatedly validated, are distilled into reusable \textit{knowledge}. 
When such knowledge is further abstracted and remains stable across tasks, it forms higher-level \textit{wisdom}. 
Cognitive accumulation therefore relies not on retaining ever more context, but on enabling context to undergo structural differentiation over time: 
short-term experience supports immediate decisions; 
relatively stable knowledge preserves strategic consistency throughout prolonged exploration; 
and further abstracted, consolidated wisdom enables transfer and reuse across tasks. 

This necessity to separate transient processing from stable state mirrors the fundamental design of multi-level-cache hierarchy in computer systems. 
It does not attempt to retain all information indefinitely, but instead relies on hierarchical structures to clearly separate short-lived, frequently accessed information from long-term, stable and reusable state under finite resources. 
ML-Master 2.0 introduces \textbf{Hierarchical Cognitive Caching (HCC)} as its core long-horizon context management architecture. 
HCC is not a single mechanism, but a coordinated design composed of two complementary components: 
(1) \textbf{Hierarchical caching}, which provides the structural architecture for organizing context into multiple tiers according to their temporal stability and reuse value; 
and (2) \textbf{Context migration}, which dictates the governance protocol for how information is dynamically promoted, consolidated, or discarded across these tiers as exploration unfolds. 
Together, these components allow rapidly changing, high-utility context to remain close to ongoing reasoning, while progressively migrating more stable and reusable knowledge into increasingly persistent representations, thereby enabling sustained and efficient long-horizon exploration.

We evaluate ML-Master 2.0 on OpenAI's MLE-Bench under a fixed 24-hour execution budget. 
As shown in Figure~\ref{fig:performance}, we measure performance using the average medal rate, defined as the percentage of tasks where the method achieves Bronze, Silver, or Gold-level performance. 
We can see that ML-Master 2.0 achieves state-of-the-art performance across all difficulty levels. 
In particular, ML-Master 2.0 attains an overall medal rate of 56.44\%, representing a 92.7\% relative improvement over ML-Master. 
The gains are consistent across task complexities: performance on low-complexity tasks improves from 48.48\% to 75.76\%, while medium-complexity and high-complexity tasks improve from 20.18\% to 50.88\% and from 24.44\% to 42.22\%, respectively. 
These results indicate that the advantages of ML-Master 2.0 stem from its ability to accumulate, stabilize, and reuse experience over extended time scales which are precisely the conditions under which ultra-long-horizon autonomy becomes critical.

Our contributions can be summarized as follows:
\begin{itemize}[left=1em]
    \item \textbf{A Conceptual Framework for Cognitive Accumulation.} We redefine ultra-long-horizon autonomy not merely as context window expansion, but as an evolutionary process. By modeling the transition from transient \textit{experience} to validated \textit{knowledge} and abstract \textit{wisdom}, we establish a theoretical basis for agentic science, enabling agents to sustain coherent reasoning over extended scientific workflows.
    \item \textbf{The Hierarchical Cognitive Caching (HCC) Architecture.} We implement this theory via ML-Master 2.0, which incorporates a multi-tiered context management system inspired by computer systems. HCC dynamically coordinates the promotion and consolidation of information, effectively decoupling high-frequency execution feedback from long-term strategic planning to overcome context saturation, a critical challenge to agentic science.
    \item \textbf{State-of-the-Art Performance on MLE-Bench.} We demonstrate that ML-Master 2.0 achieves a 56.44\% medal rate on OpenAI's MLE-Bench. These results empirically validate that structured cognitive accumulation is the critical enabler for autonomous agents to master the long-horizon trial-and-error loops characteristic of real-world scientific research.
\end{itemize}

\section{Related Work}
\subsection{Context Management}
Context management is a fundamental challenge for large language model (LLM)-based agents operating under finite context windows and long-horizon task requirements. Early approaches primarily focus on extending effective context length through architectural or system-level mechanisms, such as hierarchical context buffering and external memory paging. Representative systems such as MemGPT~\cite{packer2023memgpt} adopt an operating-system-inspired design that separates active context from external memory, enabling agents to page information in and out via explicit memory operations, often implemented through summarization or compression mechanisms. Subsequent hierarchical memory systems, including HiAgent~\cite{hu2025hiagent}, G-Memory~\cite{Zhang2025gmemory}, and graph-based retrieval frameworks such as HippoRAG~\cite{gutierrez2025hipporag}, further organize contextual information into multiple layers, allowing agents to retrieve high-level abstractions while preserving access to low-level details.

While these methods demonstrate the effectiveness of hierarchical storage and retrieval, they largely emphasize resource allocation aspects of context management, focusing on where information is stored and how it is retrieved. Memory promotion or summarization is typically applied in a heuristic manner, without explicitly modeling how execution experience evolves over time or how raw interaction traces should be selectively retained or discarded across different stages of task execution~\cite{hu2025memory,zhang2025context}.

Another line of work approaches context management from an experience-driven perspective, where execution trajectories are accumulated and abstracted to guide future behavior. Methods such as Reflexion~\cite{shinn2023reflexion}, Memento~\cite{zhou2025memento}, and ReasoningBank~\cite{ouyang2025reasoningbank} transform raw trajectories or reasoning traces into reusable feedback, cases, or strategy items. Related approaches further abstract experience into transferable artifacts, including thought templates (Buffer of Thoughts~\cite{yang2024buffer}), abstract workflows (AWM~\cite{wang2024agentworkflowmemory}), and self-evolving memory mechanisms (Evo-Memory~\cite{wei2025evo}).
 These studies highlight the importance of converting raw interaction history into compact and reusable representations, but they typically operate with flat or loosely structured memory stores and lack explicit mechanisms for regulating memory growth and lifecycle~\cite{hu2025memory}.

Taken together, existing work on context management has largely explored hierarchical organization and experiential abstraction as separate design dimensions. There remains limited investigation into frameworks that jointly regulate how short-term working context, accumulated execution experience, and abstracted memory interact within a unified control process. In particular, the absence of structured policies governing when raw interaction traces should be accumulated, promoted, or evicted constrains the ability of context management to simultaneously support scalable execution and continual adaptation~\cite{hu2025memory}.

\subsection{Autonomous Machine Learning}

The transition to autonomous machine learning necessitates managing complex, end-to-end workflows. Early benchmarks like MLAgentBench~\cite{huang2023mlagentbench} and platforms like OpenHands~\cite{wang2025openhands} established the need for tight execution feedback loops. To address extended horizons, methods such as AIDE~\cite{jiang2025aide} and R\&D-Agent~\cite{yang2025rdagent} employ iterative refinement and distinct planning phases. However, these approaches typically manage context through linear aggregation or summarization, failing to structurally distinguish between transient execution details and the stable strategic insights required for ultra-long-horizon autonomy.

To move beyond single-trajectory limitations, recent works incorporate search and evolutionary strategies. AIRA~\cite{toledo2025aira}, AutoMLGen~\cite{du2025automlgen}, and FM Agent~\cite{li2025fmagent} introduce mechanisms for cross-branch knowledge sharing and island-based evolution. While these methods enable information transfer, they generally treat "knowledge" as a homogeneous entity. They lack the cognitive differentiation necessary to distill raw \textit{experience} into reusable \textit{wisdom}, limiting their ability to sustain focus over prolonged scientific discovery.

Other approaches focus on optimizing search efficiency through IDE integration~\cite{sahney2025operand}, learned scoring models~\cite{Kulibaba2025kompeteai}, or targeted web-based initialization~\cite{nam2025mlestar}. Despite improving convergence, these methods do not explicitly address the architectural separation of processing and state. In contrast, our Hierarchical Cognitive Caching (HCC) mirrors computer memory hierarchies to dynamically govern the promotion and consolidation of context, ensuring efficient long-term exploration without saturation.

\section{ML-Master 2.0: ultra-long-horizon MLE agent with cognitive accumulation}

\subsection{Problem Formulation}
\label{sec:problem_formulation}
We formulate the interaction between the agent and the machine learning environment as a discrete event sequence $\mathcal{E}_t \triangleq\{e_0, e_1, \dots, e_{t}\}$,  where the subscript $t$ denotes chronological order (we also use $\mathcal{E}_{i:j} \triangleq (e_i,e_{i+1},\dots,e_j)$ to denote the ordered events from step $i$ to $j$). We partition the event space into environment-originated events $\mathcal{U}$ (e.g., task descriptions, user instructions, execution feedback) and agent-originated events $\mathcal{A}$ (e.g., code patches, commands, plans), and assume an alternating structure such that $e_{2k}\in\mathcal{U},e_{2k+1}\in\mathcal{A}$. Let \(g(\cdot)\) denote the context construction function that maps an interaction history to the model input context, and let \(\pi\) denote the agent policy. At any agent step ($t=2k+1$), the agent maintains a context $C_{t-1}=g(\mathcal{E}_{t-1})$ which is constructed from the interaction history $\mathcal{E}_{t-1}$, and produces an action event $e_t \sim \pi(\cdot \mid C_{t-1}), e_t\in\mathcal{A}$. The environment $\mathcal{S}$ then executes the agent’s action and produces a subsequent environment event $e_{t+1} \sim \mathcal{S}(\cdot \mid e_t), e_{t+1}\in\mathcal{U}$. For a given machine learning task $\tau_n\in \mathcal{T}$, the agent interacts with the environment until termination at the step limit $t_{\max}$, after which the final solution code $I^* = h(\mathcal{E}_{t_{\max}})$
 is obtained by applying the extraction function \(h(\cdot)\) to the terminal interaction history. The objective of the agent is to obtain a solution \(I^*\) that optimizes the task-specific evaluation metric $F(\cdot)$.

We further introduce a phase-level temporal structure induced by hierarchical research planning.
Specifically, the agent periodically proposes a hierarchical research plan, whose execution defines
a contiguous exploration phase. Let ${T}_p = \{t_0, t_1, \dots, t_p\}$ denote the set of phase
boundary time steps, where $t_{p-1}$ and $t_p$ marks the beginning and completion of the $p$-th research plan.
The interaction interval $[t_{p-1}, t_p)$ therefore corresponds to one coherent exploration phase,
typically consisting of multiple parallel implementation trajectories.

\textbf{Challenge.} In long-horizon MLE tasks, $\mathcal{E}_t$ grows rapidly due to repeated trial-and-error, tool logs, and iterative debugging. Naively setting $g(\cdot)$ to “concatenate the most recent events” causes context saturation, degrading strategic coherence and preventing accumulation of reusable expertise over tens of hours. This motivates a context management that (i) preserves high-fidelity short-term traces for immediate execution, while (ii) progressively distills stable knowledge and reusable wisdom for long-horizon planning and transfer.

\subsection{Agent Overview}
ML-Master 2.0 is an autonomous agent designed to tackle long-horizon machine learning tasks through cognitive accumulation. 
Building upon ML-Master~\cite{liu2025ml}, which optimizes exploration through rule-based MCTS, ML-Master 2.0 transitions to an agent-centric loop that explicitly manages and evolves its cognitive state over time.

\begin{figure}[t]      
    \centering            
    \includegraphics[width=1.0\linewidth]{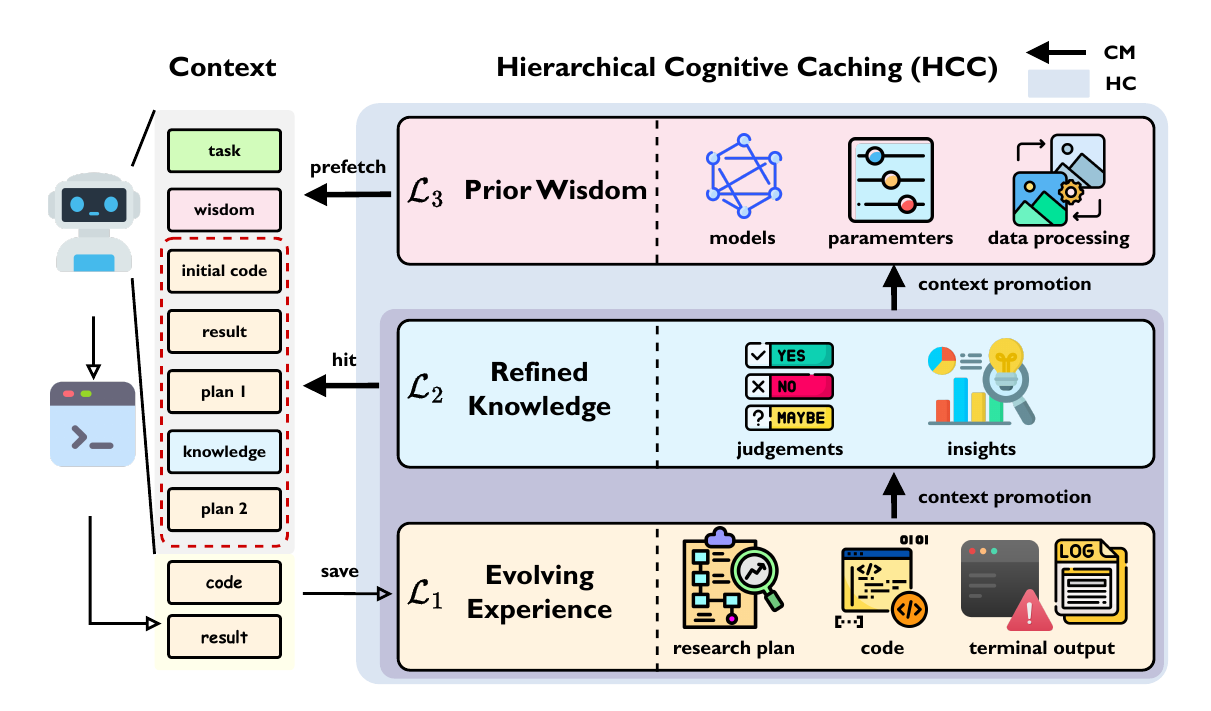} 
    \caption{The ML-Master 2.0 Framework for Ultra-Long-Horizon Autonomous MLE via Cognitive Accumulation. HC and CM represent hierarchical caching and context migration respectively. } 
    \label{fig:pipeline} 
\end{figure}

As shown in \Cref{fig:pipeline}, the core of ML-Master 2.0 is the \textbf{Hierarchical Cognitive Caching} (HCC), a coordinated design composed of two complementary key components:
(1) \textbf{Hierarchical caching}, which provides the structural architecture for organizing context into multiple tiers according to their temporal stability and reuse value,
and (2) \textbf{Context migration}, which dictates the governance protocol for how information is dynamically promoted, consolidated, or discarded across these tiers as exploration unfolds. 
For a given machine learning task, the agent first retrieves relevant prior wisdom to construct an initial context and tries to generate a buggy-free initial code through interaction with the environment. Afterwards, it proposes a hierarchical research plan, which consists of $m$ distinct exploration directions, each containing $q$ concrete implementation suggestions.  
The agent then executes these suggestions in parallel.
In terms of the temporal notation defined in \Cref{sec:problem_formulation}, this process corresponds to the event subsequence $\mathcal{E}_{t_{p-1}:t_p}$, where the interval $[t_{p-1}, t_p)$ encapsulates the cumulative interaction steps generated across all exploration directions.
This process involves generating code, interacting with the environment, and refining its approach based on the feedback received. After completion of all directions in the plan, the agent reaches a phase boundary $t_p$, consolidates the entire phase, and proposes the next research plan. This process repeats until task completion or a predefined time budget is reached.

Overall, HCC operationalizes our cognitive accumulation framework by structurally separating transient experience from stable knowledge and reusable wisdom, enabling ML-Master 2.0 to sustain coherent long-horizon exploration without being overwhelmed by execution details.

\subsection{Hierarchical Caching}
To reconcile the limited context window with the massive information generated during long-horizon exploration, we introduce a three-level hierarchical cache: $\mathcal{L}_1 \rightarrow \mathcal{L}_2 \rightarrow \mathcal{L}_3$  corresponding to Evolving Experience, Refined Knowledge and Prior Wisdom. The key idea is to separate transient context from stable cognitive state. Raw traces are kept only as needed, then promoted into compact summaries that preserve strategy, and finally distilled into reusable and transferable wisdom across tasks. Each cache layers serves as distinct role at different temporal scales. This explicit separation allows rapidly  changing signals to remain close to the active loop, while progressively consolidating stable, reusable cognition into persistent memory.

\subsubsection{$\mathcal{L}_{1}$ Cache: Evolving Experience}
Evolving Experience keeps high-fidelity execution traces required for immediate reasoning, including the current research plan, code patches and terminal outputs (e.g., error messages, metric logs).
This layer acts as the agent's \textit{working memory}, enabling precise debugging and execution-aware decision-making.
Specifically, we define
the set of plan-boundary events up to the current phase as $\mathcal{P}_{p-1} \triangleq \{ e_{t_r} \}_{r=0}^{p-1}$, and then the evolving experience cache at any time step $t \in [t_{p-1}, t_p)$ is $\mathcal{L}_1(t) = \mathcal{E}_{t_0-1}\cup\mathcal{P}_{p-1}\cup\mathcal{E}_{t_{p-1}+1:t}$. Here $\mathcal{E}_{t_0-1}$ contains all events that occur before reaching the initial code and its results, $\mathcal{P}_{p-1}$ stores all high-level plan proposals generated at phase boundaries, and $\mathcal{E}_{t_{p-1}:t}$ records the raw execution traces of the currently active plan after it is proposed. By retaining raw traces only for active execution, $\mathcal{L}_1$ preserves fidelity for immediate reasoning while preventing context saturation in long-horizon exploration.

\subsubsection{$\mathcal{L}_{2}$ Cache: Refined Knowledge}
Refined knowledge stores intermediate stabilized cognition distilled from completed exploration phases, such as key judgments (e.g., feature X is harmful),
experimental insights (e.g., CV leakage observed under split Y),
and condensed progress summaries that preserve decision rationale, while removing verbose execution details.
This layer serves as the agent's \textit{mid-term strategic memory}, maintaining coherence across iterative trial-and-error. Mathematically, define $\kappa_{i:j}$ as the compact knowledge summary from the raw event segment ${\mathcal{E}_{i:j}}$ which is obtained via a phase-level context promotion operator (see \Cref{sec:context_promotion} for details), then the $\mathcal{L}_2$ cache at any time step $t \in [t_{p-1}, t_p)$ is $\mathcal{L}_2(t) = \{\kappa_{t_{p-1}+1:t_p-1}\}_{r=1}^{{p-1}}.$ This enables long-horizon planning by allowing the agent to revisit validated decisions and insights without carrying verbose execution logs, thereby stabilizing strategic reasoning across tens of hours.

\subsubsection{$\mathcal{L}_{3}$ Cache: Prior Wisdom}
Prior Wisdom stores task-agnostic, transferable strategies distilled from previously solved machine learning tasks, such as robust model templates, reusable preprocessing pipelines, and stable hyperparameter priors. This cache serves as the agent’s long-term memory and enables warm-starting and cross-task transfer. 
Specifically, let $d_n$ denote the compact descriptor of a past task $\tau_n$, which is a high-level textual task summary generated by an LLM (see \Cref{app:prompts_prefetch} for the prompt), and let $E(\cdot)$ be a semantic embedding function.
We store prior wisdom as a set of embedding-value pairs:$\mathcal{L}_3 \triangleq {(\mathbf{h}_n, w_n)}_{n=1}^{N}$, where $N$ is the number of stored past tasks, $\mathbf{h}_n = E(d_n)$ serves as the retrieval key, and $w_n$ is the corresponding distilled task-level  wisdom text, which is obtained via a task-level context promotion operator (see \Cref{sec:context_promotion} for details). $\mathcal{L}_3$  is persistent across tasks and is only updated upon task completion via context promotion. It enables efficient transfer across tasks and provides strong priors for the bootstrap phase, improving long-horizon autonomy under limited context budgets.

\subsection{Context Migration}
\begin{figure}[!h]      
    \centering            
    \includegraphics[width=1.0\linewidth]{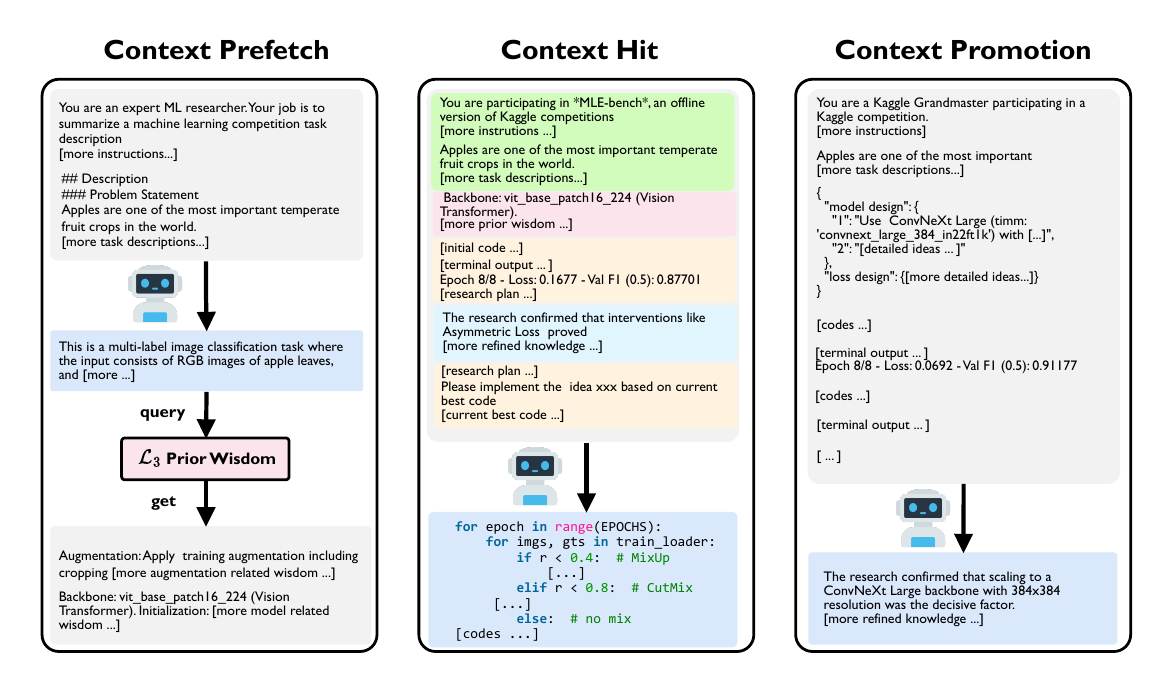} 
    \caption{An example of context migration in task \textit{plant-pathology-2021-fgvc8}.} 
    \label{fig:context_migration}  
\end{figure}

To enable genuine cognitive accumulation within the hierarchical storage, we propose a cache-like mechanism tailored for continual cognitive refinement. As shown in \Cref{fig:context_migration}, this mechanism comprises context prefetching, context hit, and context promotion.

\subsubsection{Initialization via Context Prefetching}  
The functionality of context prefetching is to let the agent construct a strong initial prior wisdom before exploration begins.
Mathematically, given the current task descriptor \(d_\tau\), we compute its embedding \(\mathbf{q}=E(d_\tau)\) and retrieve a subset of prior wisdom via a threshold-based prefetching operator $\Omega_\tau = \left\{ w_n \ \middle|\ (\mathbf{h}_n,w_n)\in\mathcal{L}_3,\  
\text{cos}(\mathbf{q},\mathbf{h}_n) > \delta \right\}$, where \(\delta\) is a similarity threshold and $\operatorname{cos}(\cdot,\cdot)$ denotes cosine similarity.  
The retrieved set \(\Omega_\tau\) represents a view of \(\mathcal{L}_3\) and leaves the cache unchanged. The initial environment event is then constructed by combining the task description, user instructions, and retrieved prior wisdom: $e_0 \triangleq \operatorname{concat}(d_\tau, u_{\text{user}}, \Omega_\tau)$. Finally, the initial agent context is $C_0 = g(\mathcal{E}_0)={e_0}$. This prefetching step ensures the agent starts with a strong and relevant context, significantly enhancing its initial understanding and enabling a more informed exploration from the very beginning.

\subsubsection{Retrieval via Context Hit}
The context constructor $g(\cdot)$ manages historical indices via a cache-like hit policy: it retrieves raw events from the evolving experience cache $\mathcal{L}_1$ when available, and otherwise falls back to compact summaries in the refined knowledge cache $\mathcal{L}_2$.
Formally, at time $t \in [t_{p-1}, t_p)$ we define
$$
\Psi_t(k)=
\begin{cases}
e_k, & e_k \in \mathcal{L}_1(t), \\
\kappa_{t_{r-1}+1:t_r-1}, & e_k \notin \mathcal{L}_1(t), e_k \in \mathcal{L}_2(t), \ k = t_{r-1}+1, \\
\varnothing, & \text{otherwise,}
\end{cases}
$$
and construct the context as
$$C_{t-1} = g(\mathcal{E}_{t-1}) = \operatorname{concat}\{ \Psi_t(k) \}_{k=0}^{t-1}.$$
This policy retrieves all plan events and active-phase traces in raw form, while representing each completed phase by a single refined unit, preventing context saturation without losing strategic continuity.

\subsubsection{Consolidation via Context Promotion}
\label{sec:context_promotion}
As parallel exploration proceeds, parts of the context become less critical for the immediate task yet remain valuable as long-term experience. The context promotion operator $P$ performs LLM-based retrospective abstraction that compresses execution traces into concise knowledge or transferable wisdom. Specifically, we decompose $P$ into two operators, $P \triangleq (P_1, P_2)$, where $P_1$ performs phase-level summarization into refined knowledge units, and $P_2$ performs task-level distillation into reusable prior wisdom.

\textbf{Phase-level promotion}. Upon completion of each exploration phase, the phase-level promotion operator compresses the raw parallel exploration trajectories into a refined knowledge unit.
Specifically, at the beginning of phase $p$, assume the agent proposes a hierarchical research plan consisting of \(m\) exploration directions, each containing \(q\) concrete implementation suggestions. Each suggestion induces an interaction trajectory:
\[
\sigma_{p,i,j} \triangleq \big(e_{a_{p,i,j}}, e_{a_{p,i,j}+1}, \dots, e_{b_{p,i,j}}\big),
\]
where $(i,j) \in \mathcal{I}_p \triangleq \{1,\dots,m\} \times \{1,\dots,q\}$ and $a_{p,i,j}$, $b_{p,i,j}$ are the start and end time steps constrained by the phase boundaries $t_{p-1} < a_{p,i,j} \le b_{p,i,j} < t_p, \forall (i,j)\in\mathcal{I}_p$. 
The phase-level promotion operator \(P_1\) maps a set of trajectories to a textual knowledge summary via LLM-based summarization(see \Cref{app:prompts_promotion_1} for detailed prompts):
\[
\kappa_p \triangleq \kappa_{t_{r-1}+1:t_r-1} = P_1\big(\{\sigma_{p,i,j}\}_{(i,j)\in\mathcal{I}_p}\big).
\]
Then the resulting summary is written to the refined knowledge cache and the corresponding raw trajectories are removed from the evolving experience cache:
$\mathcal{L}_2 \leftarrow \mathcal{L}_2 \cup \{\kappa_p\}, 
\mathcal{L}_1 \leftarrow \mathcal{L}_1 \setminus \big\{e \mid \exists (i,j)\in\mathcal{I}_p,\ e \in \sigma_{p,i,j}\big\}.$
This process incrementally distills noisy exploration traces into compact, reusable knowledge for future exploration.

\textbf{Task-level promotion}. The task-level promotion operator distills transferable wisdom from the structured task history.
Specifically, when the agent completes a task at the maximum step limit $t_{\max}$, the task-level promotion operator $P_2$ produces a task-level wisdom representation:
$$
w_\tau \triangleq P_2\big(d_\tau, \mathcal{L}_1(t_{\max}), \mathcal{L}_2(t_{\max}), h(\mathcal{E}_{t_{\max}})\big).
$$
The resulting wisdom $w_\tau$ is then embedded and stored in the prior wisdom cache:  $\mathcal{L}_3 \leftarrow \mathcal{L}_3 \cup \{(E(d_\tau), w_\tau)\}.$ 
Through this process, transient execution-level context is progressively crystallized into persistent, retrieval-ready wisdom, completing the cognitive accumulation loop.

\section{Experiment}
In this section, we conduct extensive experiments on MLE-Bench to validate the performance and effectiveness of ML-Master 2.0. MLE-Bench~\cite{chan2024mle} is a benchmark comprising 75 real-world Kaggle tasks, widely adopted for evaluating the automated machine learning capabilities of agentic systems.

\subsection{Experiment Setup}
\label{sec: experiment_setup}

\noindent \textbf{Environment and Settings.}
In our experiments, each agent is equipped with 36 AMD EPYC vCPUs and two NVIDIA GeForce RTX 4090 GPU. Every four task shares
1008GB memory and 1TB SSD to produce submissions and any intermediate files. The total time of a task is set to 24 hours. Overall, our testing environment is almost the same as the one required by MLE-Bench. In our experiment, we use Deepseek-V3.2-Speciale~\cite{liu2025deepseek} as the main backbone language model for coding and researching and Deepseek-V3.2 with thinking is sparingly used for context promotion. To build up a prior wisdom quickly, we use 407 kaggle competitions(with those in MLE-Bench excluded) as a warm up dataset. 

\noindent \textbf{Baselines.}
To provide a comprehensive comparison, we compare ML-ACE with both proprietary LLM-based methods and open-source LLM-based methods. Proprietary LLM-based methods includes OpenHands~\cite{wang2025openhands}, MLAB~\cite{huang2023mlagentbench}, AIDE~\cite{jiang2025aide}, R\&D-Agent~\cite{yang2025rdagent}, AIRA-dojo~\cite{toledo2025aira}, FM Agent~\cite{li2025fmagent} and MLE-STAR~\cite{nam2025mlestar}, etc. Open-source LLM-based methods includes ML-Master~\cite{liu2025ml} and AutoMLGen~\cite{du2025automlgen}. Due to the expensive cost of running complete MLE-Bench, we use the results reported by MLE-Bench as baselines.

\subsection{Main Results}
\label{sec:main_results}

\begin{table*}[!h]
\centering
\caption{ML-Master 2.0 outperforms all baselines on the main evaluation dimensions (percentage of achieving any medals across different machine learning task complexity levels) defined by MLE-Bench. Valid, Median+, Silver+ and Gold indicate the percentage of submissions with valid, above median score, silver medal or better and gold medal. Results are averaged over 3 runs with different random seeds and reported as mean $\pm$ one standard error of the mean (SEM). * denotes a incomplete report and the results are computed by padding incomplete seeds with failing scores. Best performances are marked in \textbf{bold}.}
\label{tab:main_results_simple}
\resizebox{\textwidth}{!}{%
\begin{tabular}{l cccc cccc}
\toprule
\multirow{2}{*}{\textbf{Agent}} & \multicolumn{4}{c}{\textbf{Medal rate in different complexity}} & \multicolumn{4}{c}{\textbf{Other evaluation dimensions}} \\
\cmidrule(lr){2-5} \cmidrule(lr){6-9}
 & \textbf{Low (\%)} & \textbf{Medium (\%)} & \textbf{High (\%)} & \textbf{Avg (\%)} & \textbf{Valid (\%)} & \textbf{Median +(\%)} & \textbf{Silver+(\%)} & \textbf{Gold (\%)} \\
\midrule
\rowcolor{gray!10} \multicolumn{9}{c}{\textbf{Proprietary LLM-based Methods}} \\
\midrule
\multicolumn{8}{l}{\textbf{MLAB~\cite{huang2023mlagentbench}}} \\
\midrule
gpt-4o-24-08 & $4.6 \pm 0.9$ & $0.0 \pm 0.0$ & $0.0 \pm 0.0$ & $1.6 \pm 0.3$ & $44.3 \pm 2.6$ & $1.9 \pm 0.7$ & $0.8 \pm 0.3$ & $0.8 \pm 0.5$ \\
\midrule
\multicolumn{8}{l}{\textbf{OpenHands~\cite{wang2025openhands}}} \\
\midrule
gpt-4o-24-08 & $12.1 \pm 1.5$ & $1.8 \pm 0.9$ & $2.2 \pm 2.2$ & $4.9 \pm 0.4$ & $52.0 \pm 3.3$ & $7.1 \pm 1.7$ &$4.0 \pm 1.0$ &$2.7 \pm 1.1$ \\
\midrule
\multicolumn{8}{l}{\textbf{AIDE~\cite{jiang2025aide}}} \\
\midrule
o1-preview & $35.9 \pm 1.9$ & $8.5 \pm 0.4$ & $11.7 \pm 1.3$ & $17.1 \pm 0.6$ & $82.8 \pm 1.1$ & $29.4 \pm 1.3$ &$13.5 \pm 0.7$ &$9.4 \pm 0.8$ \\
\midrule
\multicolumn{8}{l}{\textbf{R\&D-Agent~\cite{yang2025rdagent}}} \\
\midrule
gpt-5 & $ 68.2 \pm 2.6 $ & $21.1 \pm 1.5$ & $22.2 \pm 2.2$ & $35.1 \pm 0.4$ & $53.3 \pm 0.0 $ & $40.4 \pm 0.9$ & $28.4 \pm 1.6 $&$16.4 \pm 0.9$ \\
\midrule
\multicolumn{8}{l}{\textbf{AIRA-dojo~\cite{toledo2025aira}}} \\
\midrule
o3 & $55.0 \pm 1.5$ & $22.0 \pm 1.2 $ & $21.7 \pm 1.1$ & $31.6 \pm 0.8$ & $\mathbf{97.5 \pm 0.3 }$ & $45.5 \pm 0.8$ &$25.9 \pm 0.8$ &$17.3 \pm 0.4$ \\
\midrule 
\multicolumn{8}{l}{\textbf{FM Agent~\cite{li2025fmagent}}} \\
\midrule
Gemini-2.5-Pro & $62.1 \pm 1.5$ & $36.8 \pm 1.5$ & $33.3 \pm 0.0$ & $43.6 \pm 0.9$ & $96.9 \pm 1.2 $ & $51.6 \pm 1.2$ &$35.1 \pm 1.2$ & $\mathbf{22.7 \pm 0.8}$ \\
\midrule
\multicolumn{8}{l}{\textbf{MLE-STAR-PRO-1.5~\cite{nam2025mlestar}}} \\
\midrule
Gemini-2.5-Pro & $68.2 \pm 2.6$ & $34.2 \pm 1.5$ & $33.3 \pm 0.0$ & $44.0 \pm 1.3$ & $93.8 \pm 0.4 $ & $52.9 \pm 1.6$ &$30.2 \pm 2.9$ & $19.1 \pm 1.8$ \\
\midrule
\multicolumn{8}{l}{\textbf{Thesis}} \\
\midrule
gpt-5-codex & $65.2 \pm 1.5$ & $45.6 \pm 7.2$ & $31.1 \pm 2.2$ & $48.4 \pm 3.6$ & $90.2 \pm 2.4 $ & $56.0 \pm 2.8$ &$32.9 \pm 5.1$ & $20.0 \pm 3.4$ \\
\midrule
\multicolumn{8}{l}{\textbf{Leeroo*}} \\
\midrule
Gemini-3-pro-preview & $68.2 \pm 2.6 $ & $44.7 \pm 1.5$ & $40.0 \pm 0.0$ & $50.7 \pm 1.3$ & $50.7 \pm 1.3 $ & $50.7 \pm 1.3$ &$36.4 \pm 1.6$ & $21.3 \pm 2.0$ \\
\midrule
\rowcolor{gray!10} \multicolumn{9}{c}{\textbf{Open-source LLM-based Methods}} \\
\midrule
\multicolumn{9}{l}{\textbf{ML-Master}} \\
\midrule
 Deepseek-R1  & ${48.5 \pm 1.5}$  
 & ${20.2 \pm 2.3}$  
 & ${24.4 \pm 2.2}$  
 & ${29.3 \pm 0.8}$  
 & ${93.3 \pm 1.3}$  
 & ${44.9 \pm 1.2}$  
 & ${25.0 \pm 0.9}$  
 & ${17.3 \pm 0.8}$ \\
\midrule
\multicolumn{9}{l}{\textbf{ML-Master 2.0 (ours)}} \\
\midrule
\rowcolor{blue!10} Deepseek-V3.2-Speciale & $\mathbf{75.8 \pm 1.5}$ & $\mathbf{50.9 \pm 3.5 }$ & $\mathbf{42.2 \pm 2.2}$ & $\mathbf{56.4 \pm 2.5}$ & $95.6 \pm 1.2$ & $\mathbf{63.1 \pm 1.2}$ & $\mathbf{45.3 \pm 2.0}$&$19.6 \pm 0.9$ \\
\bottomrule
\end{tabular}%
}
\end{table*}

\noindent \textbf{ML-ACE achieves state-of-the-art performance across MLE-Bench.} As shown in \Cref{tab:main_results_simple}, ML-ACE achieves the highest average medal rate of 56.4\% among all evaluated methods, significantly outperforming the previous best proprietary LLM-based methods method with a  relative improvement of 11.2\%(increasing from 50.7\% to 56.4\%).

\noindent \textbf{ML-Master 2.0 demonstrates superior versatility across all difficulty levels.} Unlike other agents that suffer significant performance drops in complex scenarios, ML-Master 2.0 ranks first in medal rates across all complexity settings—achieving 75.8\% in Low, 50.9\% in Medium, and 42.2\% in High complexity tasks—proving its capability to handle diverse challenges effectively.

\noindent \textbf{ML-Master 2.0 exhibits exceptional robustness and high-level human capability.} ML-Master 2.0 maintains an average valid submission rate of 95.6\%, which is on par with other leading approaches. Notably, it outperforms 50\% human participants in 63.1\% of tasks, which is also the highest among all methods, indicating exceptional robustness and a high performance floor.

\subsection{Abalation Study and Analysis}

\begin{table}[!h]
  \centering
  \caption{Ablation study of the HCC architecture in ML-Master 2.0 on MLE-Bench-Lite. The full implementation of ML-Master 2.0 achieves superior performance across all metrics, illustrating the synergistic design of the HCC architecture. Best performances are marked in \textbf{bold}.}
  \label{tab:hcc_ablation}
  \begin{tabular}{cccc|ccc}
    \toprule
    & \multirow{2}{*}{\textbf{$\mathcal{L}_1$(Experience)}} &\multirow{2}{*}{\textbf{$\mathcal{L}_2$(Knowledge)}} &\multirow{2}{*}{\textbf{$\mathcal{L}_3$(Wisdom)}} &
    \textbf{Valid} &
    \textbf{Above} &
    \textbf{Any}   \\
    &  &  &  & \textbf{Submission} & \textbf{Median} & \textbf{Medal}    \\ 
    \midrule
    \ding{172} & \XSolidBrush & - & \Checkmark & 54.5 & 36.4 &  22.7  \\
    \ding{173} & \Checkmark & \XSolidBrush & \Checkmark & \textbf{95.5} & \textbf{81.8} &  59.1  \\
    \ding{174} & \Checkmark  & \Checkmark & \XSolidBrush & \textbf{95.5} & 72.7 & 54.5   \\
    \midrule
    \rowcolor{blue!9}
    \ding{175} & \Checkmark & \Checkmark & \Checkmark & \textbf{95.5} & \textbf{81.8} &  \textbf{72.7}   \\
    \bottomrule
  \end{tabular}
\end{table}

\noindent \textbf{Effectiveness of Hierarchical Cognitive Caching Architecture.}
To validate the effectiveness of the proposed Hierarchical Cognitive Caching architecture, we evaluate the individual contributions of each hierarchy-Experience ($\mathcal{L}_1$), Knowledge ($\mathcal{L}_2$), and Wisdom ($\mathcal{L}_3$)-by excluding them one at a time. Due to computational constraints, experiments are conducted on MLE-Bench-Lite with a single representative run for each configuration. Specifically: Row \ding{172} denotes the removal of Prior Experience ($\mathcal{L}_1$). In this setting, the agent lacks the capability to iteratively interact with the environment for code refinement, which consequently entails the absence of Refined Knowledge ($\mathcal{L}_2$). Row \ding{173} signifies the exclusion of Refined Knowledge ($\mathcal{L}_2$), wherein the agent is deprived of the HCC context management and storage mechanisms, resulting in all information being retained in the context in its raw format. Finally, Row \ding{174} corresponds to the omission of Wisdom ($\mathcal{L}_3$). In this scenario, the agent is unable to retrieve wisdom or leverage strategies from historically similar tasks. 

As shown in \Cref{tab:hcc_ablation}, the full implementation of ML-Master 2.0 achieves superior performance across all metrics, validating the synergistic design of the HCC architecture. The removal of Evolving Experience ($\mathcal{L}_1$) leads to a substantial deterioration in performance, with the valid submission rate plummeting to 54.5\% and the medal rate to 22.7\%. This underscores that the evovling experience's foundational role when handling ultra-long-horizon tasks.

In setting \ding{173}, the drop of the any medal rate indicates that while retaining raw context allows for average performance, the Refined Knowledge is indispensable for synthesizing the complex solutions required to reach top-tier performance.

The omission of Prior Wisdom ($\mathcal{L}_3$) in setting \ding{174} results in a degradation across both above median rate(72.7\%) and any medal rate(54.5\%). This result indicates that Prior Wisdom provides a strong initialization that substantially reduces ineffective exploration, which is critical for achieving competitive performance.

\begin{figure}[!h]      
    \centering          
    \includegraphics[width=1.0\linewidth]{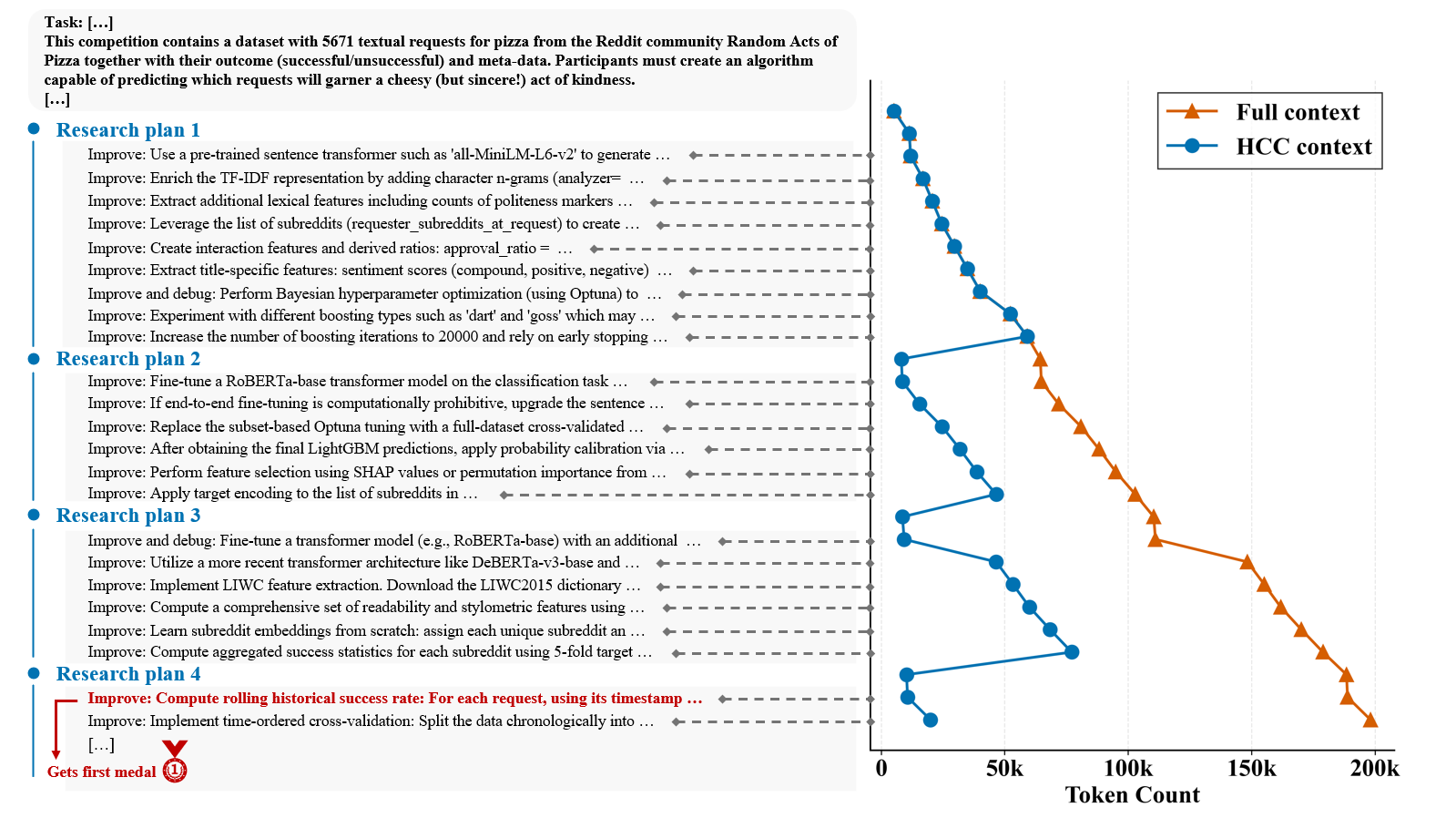} 
    \caption{The growth of context length when ML-Master 2.0 is handling the task \textit{random-acts-of-pizza}. The orange line represents the full context length while the blue line represents the context length in HCC. ML-Master 2.0 successfully limits the peak context length from more than 200k to approximately 70k tokens and secures a medal during the fourth iteration of the research plan proposal and verification.}
    \vspace{-10pt}
    \label{fig:context_length}  
\end{figure}

\noindent \textbf{Effectiveness of Context Promotion. }
\Cref{fig:context_length} illustrates the growth of context length as ML-Master 2.0 addresses a specific ultra-long-horizon task under the HCC architecture. It can be observed that without intervention, context length proliferates rapidly during complex tasks, particularly when debugging is introduced additionally to hanlde errors in terminal outputs. Conversely, with the help of the HCC architecture, ML-Master 2.0 effectively limits the peak context length from more than 200k to approximately 70k tokens while still retaining critical insights from prior failed attempts. Consequently, it successfully secures a medal during the fourth iteration of the research plan proposal and verification.

\begin{wrapfigure}{r}{0.4\textwidth}
    \centering
    \vspace{-20pt}
    \includegraphics[width=1\linewidth]{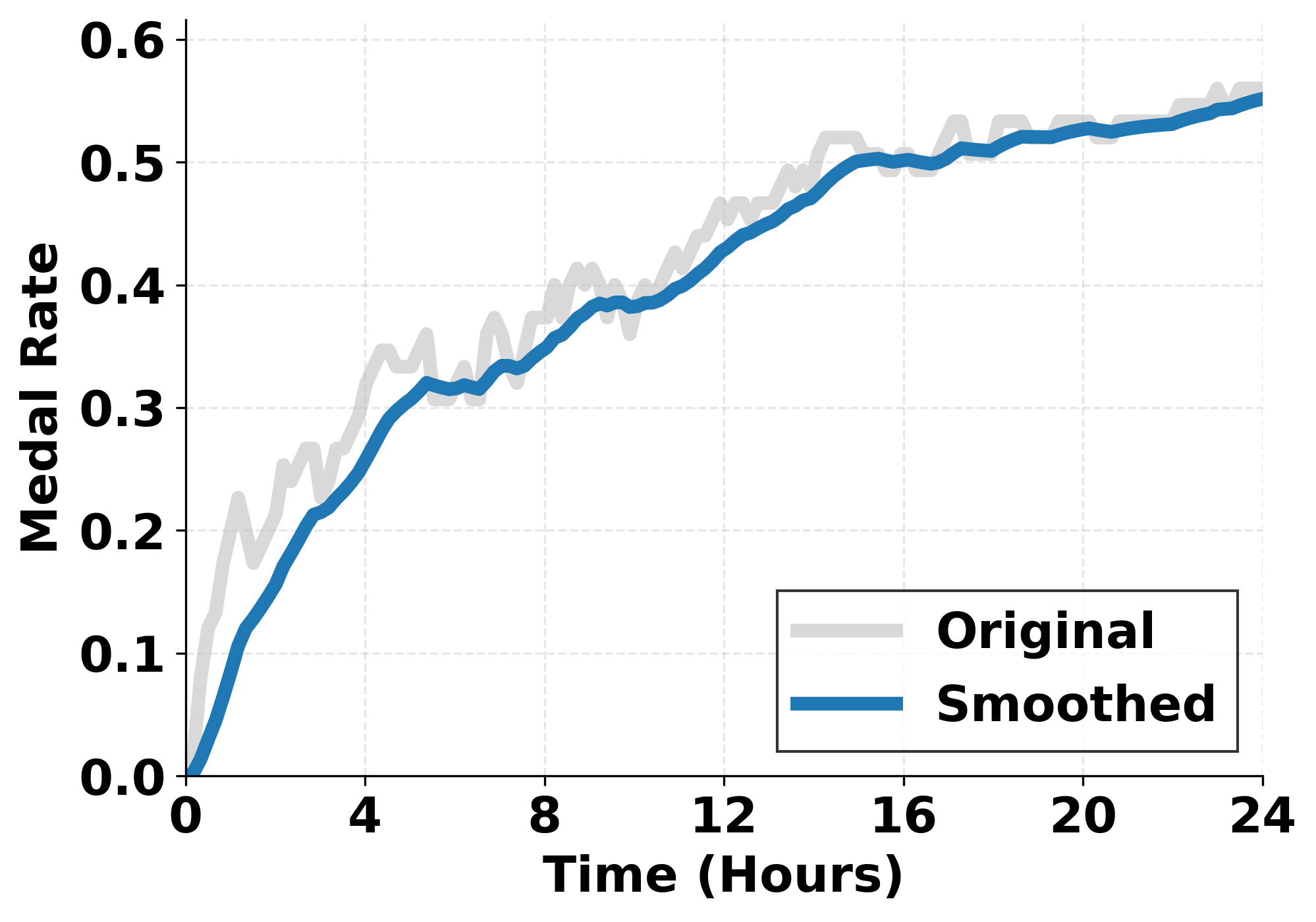}
    \caption{ML-Master 2.0 continues improving its solution over time.}
    \label{fig:time_scale}
    \vspace{-45pt}
\end{wrapfigure}

\noindent \textbf{Peformance over time. } 
ML-Master 2.0 continues improving its solution over time. In \Cref{fig:time_scale}, we show how the solution given by ML-Master 2.0 evolves over time. The vertical axis represents average medal rate while the horizontal axis represents the iteration time. As the iteration time increases, ML-Master 2.0 produces increasingly better solutions, demonstrating the effectiveness of its HCC architecture.

\section{Conclusions}
In this work, we have presented ML-Master 2.0, a framework that addresses the critical bottleneck of ultra-long-horizon autonomy in agentic science. By shifting the paradigm of context management from linear retention to cognitive accumulation, our Hierarchical Cognitive Caching (HCC) architecture enables agents to dynamically distill transient execution experiences into refined knowledge and reusable prior wisdom. This structural differentiation allows the agent to sustain strategic coherence over extended experimental cycles without being overwhelmed by the exponential growth of interaction details.
Empirical evaluations on OpenAI's MLE-Bench demonstrate that ML-Master 2.0 achieves a state-of-the-art medal rate of 56.44\%, significantly outperforming existing open-source and closed-source baselines. These results validate that the ability to evolve context is essential for mastering the high-dimensional, delayed-feedback environments characteristic of real-world scientific research. Ultimately, ML-Master 2.0 establishes a foundational paradigm for agentic science, offering a scalable blueprint for autonomous agents capable of orchestrating the full lifecycle of scientific discovery beyond human-precedent complexities.

\newpage
\bibliography{main}

\newpage
\appendix
\renewcommand{\sectionautorefname}{Appendix}
\section{Prompts for ML-Master 2.0}
\label{app:prompts}
\subsection{Context Prefetching}
\label{app:prompts_prefetch}
\begin{promptbox}[Prompts for generating compact descriptor $d_n$ for task $\tau_n$]
You are an expert ML researcher. Your job is to summarize a machine learning competition task description into a single dense paragraph optimized for embedding-based retrieval.\\

Your summary must:

- Focus ONLY on essential technical details present in the description.

- Be dense, precise, and semantic-rich.

- Stay strictly below 250 tokens.

- Contain all relevant elements: task type, input format, output format, evaluation metric (and how it is computed), dataset structure and key fields/modalities, submission format, and any important constraints or rules.\\

STRICT FORMAT REQUIREMENTS:

- Output MUST be a single paragraph.

- No bullet points, no numbering, no markdown.

- No headings, no blank lines, no lists.

- No backticks or code blocks.

- No special characters other than standard punctuation.

- Do NOT introduce information that is not explicitly present in the task description.

- Do NOT explain ML concepts or provide suggestions, analysis, or background.\\

TASK DESCRIPTION:

----------------

\{task\_description\}

----------------\\

Now output ONLY the summarized single paragraph, with no explanations:
\end{promptbox}

\newpage
\subsection{Coding prompts}
\label{app:prompts_drafting}
\begin{promptbox}[Prompts for drafting an initial code]
You are a Kaggle grandmaster attending a competition. In order to win this competition, you need to come up with an excellent and creative plan for a solution and then implement this solution in Python. We will now provide a description of the task.\\

\# Task description

\{task\_description\}\\

\# Instructions

\#\# Response format

Your response should be a brief outline/sketch of your proposed solution in natural language (3-5 sentences), followed by a single markdown code block (wrapped in ```) which implements this solution and prints out the evaluation metric. There should be no additional headings or text in your response. Just natural language text followed by a newline and then the markdown code block.\\

\#\# Solution sketch guideline

- - The solution sketch should be 3-5 sentences.

- - Propose an evaluation metric that is reasonable for this task.

- - Don't suggest to do EDA.

- - The data is already prepared and available in the `./input` directory. There is no need to unzip any files.\\

\#\# Implementation guideline

- The code must not only implement the proposed solution but also **print the evaluation metric computed on a hold-out validation set**. **Without this metric, the solution cannot be evaluated, rendering the entire code invalid.**,

- **AND MOST IMPORTANTLY SAVE PREDICTIONS ON THE PROVIDED UNLABELED TEST DATA IN A `submission.csv` FILE IN THE ./submission/ DIRECTORY.**
- The code should be a single-file python program that is self-contained and can be executed as-is.

- No parts of the code should be skipped, don't terminate the before finishing the script.

- Your response should only contain a single code block.

- All the provided input data is stored in "./input" directory.

- **You MUST submit predictions on the provided unlabeled test data in a `submission.csv` file** file in the "./working" directory as described in the task description** This is extremely important since this file is used for grading/evaluation. DO NOT FORGET THE submission.csv file!

- You can also use the "./working" directory to store any temporary files that your code needs to create.

- REMEMBER THE ./submission/submission.csv FILE!!!!! The correct directory is important too.

- If you use `DataLoader`, you need to increase the parameter `num\_workers` to speed up the training process.\\

\#\# Installed Packages

Your solution can use any relevant machine learning packages such as: `pandas`, `statsmodels`, `torch-geometric`, `bayesian-optimization`, `torch`, `xgboost`, `spacy`, `timm`, `scikit-learn`, `transformers`, `nltk`, `lightGBM`, `numpy`, `torchvision`. Feel free to use any other packages too (all packages are already installed!). For neural networks we suggest using PyTorch rather than TensorFlow.\\

\# Data preview

\{data\_preview\}\\

\# External knowledge

Here are some information about the data loading and preprocessing and model selection and the model training from kaggle experts. Your final code must follow and use these knowledge.\\

\#\# Data loading and preprocessing

\{data\_knowledge\}

\#\# Model selection and model training

\{model\_knowledge\}
    
\end{promptbox}

\newpage
\label{app:prompts_debugging}
\begin{promptbox}[Prompts for debugging]
You are a Kaggle grandmaster attending a competition. Your previous solution had a bug and/or did not produce a submission.csv, or the generated submission.csv was in an incorrect format,so based on the information below, you should revise it in order to fix this. Your response should be an implementation outline in natural language, followed by a single markdown code block which implements the bugfix/solution.\\

\# Task description

\{task\_description\}\\

\# Instructions

\#\# Response format

Your response should be a brief outline/sketch of your proposed solution in natural language (3-5 sentences), followed by a single markdown code block (wrapped in ```) which implements this solution and prints out the evaluation metric. There should be no additional headings or text in your response. Just natural language text followed by a newline and then the markdown code block.\\

\#\# Bugfix improvement sketch guideline

- - You should write a brief natural language description (3-5 sentences) of how the issue in the previous implementation can be fixed.

- - Don't suggest to do EDA.

- - You should keep the core method of the machine learning code same. Do not change the machine learning method and just fix the code.

- - If the code failed because of missing library, try to avoid using the missing library. Do not try to install the missing library.

- - All packages have been installed. You are not allowed to install anything with pip or conda. If something is missing, try another way instead of installing a package.\\

\#\# Implementation guideline

- The code must not only implement the proposed solution but also **print the evaluation metric computed on a hold-out validation set**. **Without this metric, the solution cannot be evaluated, rendering the entire code invalid.**,

- **AND MOST IMPORTANTLY SAVE PREDICTIONS ON THE PROVIDED UNLABELED TEST DATA IN A `submission.csv` FILE IN THE ./submission/ DIRECTORY.**

- The code should be a single-file python program that is self-contained and can be executed as-is.

- No parts of the code should be skipped, don't terminate the before finishing the script.

- Your response should only contain a single code block.

- All the provided input data is stored in "./input" directory.

- **You MUST submit predictions on the provided unlabeled test data in a `submission.csv` file** file in the "./working" directory as described in the task description** This is extremely important since this file is used for grading/evaluation. DO NOT FORGET THE submission.csv file!

- You can also use the "./working" directory to store any temporary files that your code needs to create.

- REMEMBER THE ./submission/submission.csv FILE!!!!! The correct directory is important too.

- If you use `DataLoader`, you need to increase the parameter `num\_workers` to speed up the training process.\\

\# Data preview

\{data\_preview\}\\

\# Previous (buggy) implementation

\{buggy\_code\}\\

\# Execution output

\{terminal\_output\}
\end{promptbox}

\newpage
\label{app:prompts_reseach}
\begin{promptbox}[Prompts for generating a research plan]
You are a Kaggle Grandmaster participating in a Kaggle competition. I will provide you with the following information in order: (1) the competition description, (2) a preview of the dataset format, and (3) the initial code and the current best-performing code. (4) your memory including your previous tried research plans and corresponding summarized results\\

Competition Information:

\{task\_description\}\\

Dataset Preview:

\{data\_preview\}\\

Initial code:

\{initial\_code\}\\

Current Best Code:

\{best\_code\}\\

Memory(your previous tried research plans and corresponding summarized results):

\{memory\}\\

Based on the information above and the best code provided, identify at least 3 major directions where the solution can be improved to potentially achieve better performance. For each major area, propose some highly practical and feasible detailed suggestions.

Do not suggest ensembling methods. 

Do not suggest k-cross validation with k larger than 5.

You suggestions should not have any ambiguity. Avoid using `e.g.` and `or` in your answer.

Your response must strictly follow a JSON format. `major direction k` and `Your detailed suggestion k` should be replaced with your concrete answer. \\

Below is an example:

\{

    "major direction 1": \{
    
        "1": "Your detailed and specific suggestion 1",
        
        "2": "Your detailed and specific suggestion 2"
        
    \},
    
    "major direction 2": \{
    
        "1": "Your detailed and specific suggestion 1",
        
        "2": "Your detailed and specific suggestion 2",
        
    \},
    
    "major direction 3": \{
    
        "1": "Your detailed and specific suggestion 1",
        
        "2": "Your detailed and specific suggestion 2",
        
    \}
    
\}
\end{promptbox}

\newpage
\label{app:prompts_improving}
\begin{promptbox}[Prompts for improving]
You are a Kaggle grandmaster attending a competition. You are provided with previous memory including previously developed solutions and an creative idea. You need to implement this idea on top of (or building upon) the previously developed solution and memory.\\

\# Task description

Here is the original kaggle task description.

\{task\_description\}\\

\# Instructions

Here is the instruction about response format and implementation.

\#\# Response format

Your response should be a brief outline/sketch of the solution in natural language (3-5 sentences), followed by a single markdown code block (wrapped in ```) which implements this solution and prints out the evaluation metric. There should be no additional headings or text in your response. Just natural language text followed by a newline and then the markdown code block.\\

\#\# Solution improvement sketch guideline

- - The solution sketch should be a brief natural language description of how you improved the previous solution.

- - The solution sketch should be 3-5 sentences.

- - Don't do EDA.

- - All packages have been installed. You are not allowed to install anything with pip or conda. If something is missing, try another way instead of installing a package.\\

\#\# Implementation guideline

- The code must not only implement the creative idea but also **print the evaluation metric computed on a hold-out validation set**. **Without this metric, the solution cannot be evaluated, rendering the entire code invalid.**,

- **AND MOST IMPORTANTLY SAVE PREDICTIONS ON THE PROVIDED UNLABELED TEST DATA IN A `submission.csv` FILE IN THE ./submission/ DIRECTORY.**

- The code should be a single-file python program that is self-contained and can be executed as-is.

- No parts of the code should be skipped, don't terminate the before finishing the script.

- Your response should only contain a single code block.

- All the provided input data is stored in "./input" directory.

- **You MUST submit predictions on the provided unlabeled test data in a `submission.csv` file** file in the "./working" directory as described in the task description** This is extremely important since this file is used for grading/evaluation. DO NOT FORGET THE submission.csv file!

- You can also use the "./working" directory to store any temporary files that your code needs to create.

- REMEMBER THE ./submission/submission.csv FILE!!!!! The correct directory is important too.

- If you use `DataLoader`, you need to increase the parameter `num\_workers` to speed up the training process.\\

\# Data preview

Here is a preview of the real structure of the data. 

\{data\_preview\}\\

\# Previous memory and solution

\{previous\_memory\_solution\}\\

\# Creative idea

This is a creative idea which may improve the performance. You need to implement this idea on top of (or building upon) the previous solution above.

Creative idea:

\{improve\_idea\}
\end{promptbox}

\newpage
\subsection{Context promotion}
\label{app:prompts_promotion_1}
\begin{promptbox}[Prompts for phase-level context promotion $P_1$]
You are a Kaggle Grandmaster with critical thinking skills participating in a high-stakes competition. I will provide you with: (1) the competition task, (2) your memory of previous attempted research plans and summarized results, and (3) your current research plan along with its raw results (code outputs, logs, etc.).\\

Competition Information:

\{task\_description\}\\

Memory:

\{memory\}\\

Current Research Plan:

\{research\_plan\}\\

Corresponding Results:

\{results\}\\

\# Your Task

Your goal is to perform a deep analysis of the current experiment and synthesize it into a strategic summary. Your summary needs to be concise but informative.
**Do not just describe what happened.** You must evaluate the **value** of the result.\\

Your output must cover two key aspects:

1.  **Execution Summary:** Concisely state whether the plan worked as intended, the performance achieved.

2.  **Strategic Insights \& Future Direction:** This is the most important part. Based on the current results AND your memory of past attempted research plans:

    * Identify **High-Potential Directions**: Which direction seems to be promising? Which direction should be amplified or iterated upon in the next step?
    
    * Identify **Dead Ends / Low-Value Paths**: Which directions are clearly ineffective or have reached a performance plateau? Explicitly advise against continuing in these specific directions to save compute resources.\\

\# Response Format

Your output should contain **only** the final analysis text. 

Do not add any explanations, comments, greetings, or extra sentences before or after the summary.

Do not wrap the answer with phrases like "Here is the summary". Output the content directly.
\end{promptbox}

\newpage
\label{app:prompts_promotion_2}
\begin{promptbox}[Prompts for phase-level context promotion $P_2$]
Here is the kaggle task description, exploration trajectories and the final high performance code of the task \{task\_name\}. \\

Based on these information. Your job is to 

1. summarize the key point about the data loading and preprocessing of the best code.

2. summarize the key point about the model selection and the model training of the best code.

Your response should be concise but not too short. Do not omit any parameters. You should make sure an code engineer can basically reproduce the code with your summarization.\\

Kaggle task description:

\{task\_description\}\\

Trajectories and final code:

\{trajectories\}\\

Your answer should follow the format below:

DATA SUMMARY:

YOUR ANSWER

MODEL SUMMARY:

YOUR ANSWER
\end{promptbox}

\end{document}